\documentclass[letterpaper, 10 pt, conference]{ieeeconf}  

\IEEEoverridecommandlockouts                              

\usepackage{amsfonts,amssymb,amsmath}
\usepackage{algorithmic}
\usepackage{algorithm}
\usepackage{array}
\usepackage[caption=false,font=normalsize,labelfont=sf,textfont=sf]{subfig}
\usepackage{textcomp}
\usepackage{stfloats}
\usepackage{url}
\usepackage{verbatim}
\usepackage{graphicx}
\usepackage{cite}
\usepackage{siunitx}
\usepackage{makecell}
\usepackage{multirow}
\usepackage{booktabs}
\usepackage{subcaption}
\newcommand\subfootnote[1]{%
  \begingroup
  \renewcommand\thefootnote{}\footnote{#1}%
  \addtocounter{footnote}{-1}%
  \endgroup
}
\graphicspath{{figures}}
\usepackage{xcolor}
\usepackage{diagbox}   
\usepackage{array}
\usepackage{siunitx} 
\usepackage[hidelinks]{hyperref}

\title{\LARGE \bf
Multifingered force-aware control for humanoid robots
}
\author{Pasquale Marra$^{1,2}$, Gabriele M. Caddeo$^{1}$, Ugo Pattacini$^{3}$ and Lorenzo Natale$^{1}$
}

\usepackage{eso-pic} 
\newcommand{\firstpagecopyright}{%
    \AddToShipoutPictureFG*{%
        \AtPageUpperLeft{%
        \hspace*{\dimexpr1in+\oddsidemargin\relax}%
        \raisebox{-3.5\baselineskip}[0pt][0pt]{%
            \begin{minipage}{\textwidth}
            \centering\footnotesize
            \textit{\textcopyright~2026 IEEE. Personal use of this material is permitted.
            Permission from IEEE must be obtained for all other uses, in any current or future media,
            including reprinting/republishing this material for advertising or promotional purposes,
            creating new collective works, for resale or redistribution to servers or lists, or reuse of
            any copyrighted component of this work in other works.}\\
            Preprint version (Mar.\ 2026). This work has been accepted for publication in ICRA 2026.
            \end{minipage}
        }
        }
    }
}

\begin{document}
\firstpagecopyright 

\twocolumn[{%
\renewcommand\twocolumn[1][]{#1}%
\maketitle
\begin{center}
    \vspace{-0.1in}
    \centering
    \captionsetup{type=figure}
    \includegraphics[width=\linewidth]{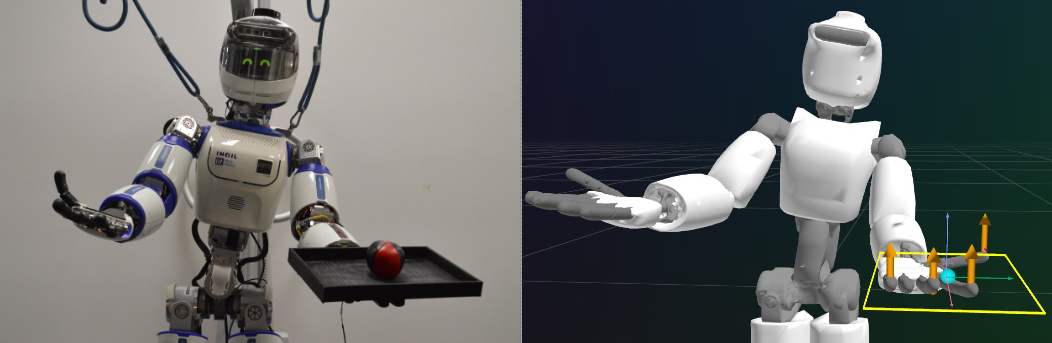}
    \captionof{figure}{Real robot (left) and its virtual
    counterpart \cite{RerunSDK} (right). ergoCub, equipped with tactile sensors, balances a tray with an unknown object by minimizing the distance between the supporting polygon center and the estimated Center of Pressure (cyan sphere), computed from normal forces (orange arrows).}
    \label{fig:first_image}
    \vspace{-0.1in}
\end{center}
}]
\subfootnote{This work was supported by the National Institute for Insurance against Accidents at Work (INAIL) project ergoCub-core.}

\subfootnote{
\\
\textsuperscript{$1$}Humanoid Sensing and Perception, Istituto Italiano di Tecnologia, Genoa, Italy. Acknowledge financial support from the
PNRR MUR project PE0000013-FAIR. \\
\textsuperscript{$2$}DIBRIS, Universit\`a di Genova, Via All'Opera Pia, 13, Genoa, Italy.,
Genoa, Italy.  \\
\textsuperscript{$3$} MESH Facility, Istituto Italiano di Tecnologia, Genoa, Italy \\

}
\thispagestyle{empty}
\pagestyle{empty}

\begin{abstract}
In this paper, we address force-aware control and force distribution in robotic platforms with multi-fingered hands. Given a target goal and force estimates from tactile sensors, we design a controller that adapts the motion of the torso, arm, wrist, and fingers, redistributing forces to maintain stable contact with objects of varying mass distribution or unstable contacts. To estimate forces, we collect a dataset of tactile signals and ground-truth force measurements using five Xela magnetic sensors interacting with indenters, and train force estimators. We then introduce a model-based control scheme that minimizes the distance between the Center of Pressure (CoP) and the centroid of the fingertips contact polygon. Since our method relies on estimated forces rather than raw tactile signals, it has the potential to be applied to any sensor capable of force estimation. We validate our framework on a balancing task with five objects, achieving a $82.7\%$ success rate, and further evaluate it in multi-object scenarios, achieving $80\%$ accuracy. Code and data can be found here \href{https://github.com/hsp-iit/multifingered-force-aware-control}{https://github.com/hsp-iit/multifingered-force-aware-control}.

\end{abstract}

\section{Introduction}
Robots meant to live alongside and assist humans must continuously refine their actions through sensory feedback \cite{doi:10.1126/science.aea2492}. While computer vision has advanced dramatically—reaching or surpassing human-level performance in some domains—tactile sensing still lags behind. Vision can substitute for some perceptual aspects of touch, but it poorly captures contact dynamics: the interaction forces are difficult to measure reliably with external cameras.
Humans routinely exploit touch for prehensile and non-prehensile manipulation: when holding a box with unknown contents or serving a tray, shifts in mass distribution are compensated by posture and hand adjustments driven by subtle force variations. Accordingly, many tactile sensors have been proposed, including finger-like, flat, and all-around designs based on magnetic \cite{magneticjamone, magnetic2}, capacitive \cite{capacitivejamali}, or vision-based principles \cite{gelslim4, digit}. Their multimodal capabilities have enabled pose \cite{caddeo2023collisionawareinhand6dobject, Bauza_2023}, shape/curvature \cite{10610028, shahidzadeh2024actexploreactivetactileexploration}, dynamic control \cite{oller2024tactiledrivennonprehensileobjectmanipulation}, and texture estimation \cite{10610274}. However, outputs differ substantially across technologies, and even within the same sensor family; therefore, manipulation methods that consume raw tactile signals often do not transfer across sensor types. 
A promising alternative is to map tactile outputs into the force domain: estimating 3D contact forces yields a sensor-agnostic representation that generalizes more robustly across devices \cite{Shahidzadeh2024FeelAnyForceEC, chen2025generalforcesensationtactile}. Yet, although force feedback is widely used in manipulation, multifingered strategies are typically limited to prehensile tasks, whereas non-prehensile approaches often rely on single-finger feedback \cite{9561350}. In this work, we introduce a model-based indirect control strategy to balance in-hand objects with varying mass distribution by distributing contact forces across a multi-fingered hand. We maximize stability by minimizing the distance between the Center of Pressure (CoP), estimated from fingertip force measurements, and the geometric center of the fingertip contact polygon.To enable force feedback, we train a lightweight network to estimate 3D forces from custom magnetic Xela sensors \cite{xelarobotics2025uscu} using a calibrated testbed with 3D-printed indenters. The model is deployed on the ergoCub \cite{ergocub} hand equipped with these fingertip sensors. A 3D-printed tray is placed on the hand and an object of unknown mass is positioned arbitrarily on it. Estimated fingertip forces are processed by a controller that computes the CoP and the desired plane rotation, then moves the fingertips to minimize the error between the CoP and the geometric center of the contact polygon, while keeping the fingertip positions on the plane. In parallel, a secondary controller supports finger control to maintain continuous contact with the tray and compensate for sources of error. This layered structure enables stable balancing while remaining agnostic to the tactile sensor type. The approach can serve as a robust low-level primitive within hierarchical frameworks or as a standalone solution for force-based in-hand balancing. We validate it by balancing an unknown object on the tray and preventing it from falling; an example is shown in Fig.~\ref{fig:first_image}.
In summary, the contributions of this work are threefold:
\begin{itemize}
    \item We present a dataset containing calibrated 3D forces, relative poses, and tactile outputs for our custom finger-like Xela sensors, collected with a dedicated testbed.
    \item We provide a comprehensive characterization of the sensor performance and limitations, offering insights into its applicability for force estimation in manipulation.
    \item We propose a multifingered control algorithm for balancing in-hand objects by acting on force distribution. The controller operates directly in the force domain, making it conceptually applicable across different tactile sensors.
\end{itemize}

\section{Related works}
Our work is closely related to manipulation control frameworks using tactile sensing. While there remains interest in solving manipulation tasks without tactile feedback \cite{10611300}, there is a growing focus on integrating tactile sensors into robotic hands control \cite{10611404, 10.3389/fnbot.2022.843267}. Early works primarily relied on single-finger feedback. For instance, in \cite{9561350}, the structural similarity index measure (SSIM) computed from TacTip images was used to control the closure of a Pisa/IIT SoftHand \cite{0278364913518998}. Similarly, in \cite{chelly2025tactilebasedforceestimationinteraction}, Xela uSkin sensors were calibrated for force estimation using an on-hand method, and validated by controlling a single Allegro hand finger \cite{AllegroHand2024}. More recent works explore multi-finger control for prehensile tasks. In \cite{10146043}, a joint-level adaptive torque controller tracked desired finger velocities, while BioTac sensors \cite{Wettels2014} were used to estimate the contact frame orientation. In \cite{17501}, grasp forces of the Allegro hand were modulated based on shear measurements from five MicroTac sensors. Other approaches estimate pseudo-forces from tactile signals: e.g., \cite{kitouni2024} employed pseudo-forces reconstructed from Xela uSkin outputs for force direction control, while \cite{deng2020grasping} used pseudo-forces derived from BioTac sensors for grasp force regulation. In \cite{lee2025trajectoryoptimizationinhandmanipulation}, fingertip forces were obtained by averaging raw outputs of magnetic tactile sensors to roll a cylinder with two fingers. Whole-body controllers have been proposed for humanoid balancing and locomotion tasks \cite{7363429, 10758235}.
These approaches similarly exploit a large number of joints and degrees of freedom to achieve stability.
However, they do not control the fingers explicitly, and therefore cannot address multifingered in-hand manipulation.
In \cite{10169079}, a tray is fixed at the end of a robotic manipulator to perform non-prehensile transportation. However, it is assumed that the object’s shape and dynamical properties are known. In contrast, we develop an upper-body controller that extends the multi-joint reasoning of whole-body approaches to also include multicontact control of the fingers of a robotic hand, while making no assumption about the object. To the best of our knowledge, this is the first work to combine an upper-body controller with multifingered force distribution for non-prehensile manipulation.

\section{Method}
In our scenario, a humanoid robot equipped with a multi-fingered hand and fingertip force sensors must balance an object with varying mass distribution, using fingertip force feedback to adjust its hand pose.

\begin{figure*}
    
    \vspace{0.5em}
    \centering
    \includegraphics[width=\textwidth]{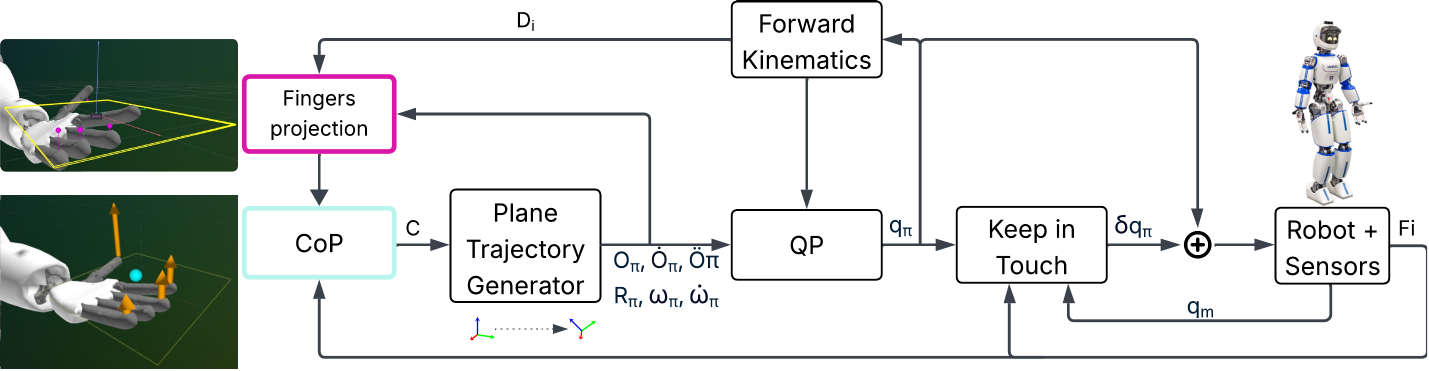}
    \caption{Overview of the control pipeline. Finger positions (magenta spheres) are projected onto the plane (yellow rectangle), and normal force components (orange arrows) are used to compute the CoP (cyan sphere) in the plane frame. From this, a new target pose for the plane is generated, and finger and joint positions are updated with a corrective term for inaccuracies.\label{fig:scheme}}
    \vskip -2.em
\end{figure*}

Let \(\mathcal{F}_\Pi = (O_{\Pi}, R_\Pi)\) be the orthonormal frame with origin in \( O_{\Pi} \in \mathbb{R}^3 \) and orientation \(R_\Pi = [b, t, n] \in SO(3)\), with \( \dot{O}_{\Pi}, \ddot{O}_{\Pi}\) denoting its linear velocity and acceleration, and \(\omega_\Pi, \dot{\omega}_\Pi \) its angular velocity and angular acceleration, respectively. Let \( \Pi \) be the plane passing through the origin \( O_{\Pi}\) with unit normal vector \( n \in \mathbb{R}^3 \). Our objective is to design a control architecture that:
\begin{itemize}
    \item[-] controls the finger positions so as to establish contact with a planar base coincident with the plane \( \Pi \);
    \item[-] employs finger force measurements to adapt the pose of the frame \(\mathcal{F}_\Pi \), and consequently the pose of the plane \( \Pi \).
\end{itemize}
In this configuration, a planar object can be placed on the fingertips, establishing a correspondence with the virtual plane \(\Pi\). This correspondence need not be exact, since the controller ensures fingertip contact with the object to compensate for misalignment. Moreover, we do not attempt to directly control the forces measured at the fingertips, as no assumptions are made about the object’s inertial properties. Fig.~\ref{fig:scheme} shows our pipeline.
This section proceeds as follow: in \ref{subs:ForceEst} we detail the characterization of the sensors and the dataset collection for force estimation;
in \ref{subs:FingerPositionsControl} we define the finger positions control; then in \ref{subs:PlanePose} we show how the fingertip force measurements are used to affect the plane \(\Pi\) pose, leveraging the concept of Center of Pressure (CoP).  
\subsection{Sensor characterization and Dataset collection}\label{subs:ForceEst}
\subsubsection{Setup}\label{subsub:setup}To characterize the sensor and collect the dataset, we use the setup shown in Fig.~\ref{fig:setup}a. A 3D-printed base secures the fingertip, while 3D-printed indenters (similar to those in \cite{9369877}, Fig.~\ref{fig:setup}b) are mounted on the end effector of the Omega3 robot \cite{omega}. An ATI Nano17 force/torque sensor is placed between the robot and the indenter to measure applied forces, which are then projected onto the fingertip frame using the known transformation between the F/T sensor and the fingertip.  For all experiments, we subtract an offset for each tactile sensor, estimated by averaging a thousand readings when the sensor is not in contact.
\subsubsection{Sensor characterization}\label{subsub:characterization}
We employ a custom version of the Xela sensors specifically designed for the ErgoCub hand (called uSCu). Each fingertip consists of five magnetic taxels that can displace in three directions (x, y, z) under applied force. Before collecting the dataset, we first characterize the sensor behavior with respect to three aspects most relevant to our scenario: \textbf{repeatability} of the measurements, \textbf{consistency} across different sensors, and the effect of \textbf{gravity} on tactile readings. 

To assess repeatability, we repeatedly press an indenter on the fingertip by controlling the robot’s position with $0.01~mm$ precision. The robot maintains contact for five seconds, rests for another five seconds, and repeats the motion. This experiment is conducted three times at reference positions corresponding to contact forces of approximately $2~N$, $4~N$, and $6~N$. Higher forces are not considered, since the fingertips are mechanically tested only up to $5~N$. While the sensor produces consistent outputs at $2~N$, performance deteriorates at higher forces due to hysteresis. Fig.~\ref{fig:setup}c illustrates this effect: the top plot shows the sequence of applied forces, while the bottom plot reports the output of one taxel, highlighting the growing influence of hysteresis with repeated touches.

\begin{figure}[t]
	\centering
	\vspace{0.5em}
    \includegraphics[width=\columnwidth]{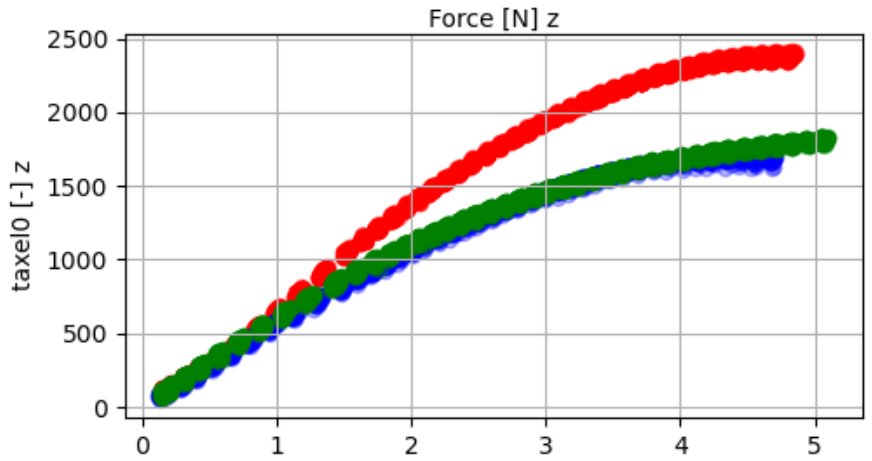}
	\caption{Consistency validation across sensors shows variability in tactile outputs under identical applied forces. \label{fig:consistency}}
    \vskip -1.8em
\end{figure}

To validate consistency, we perform a similar test across different fingertips. The robot is controlled in position along the z-axis, with the reference incremented by $0.2~mm$ at each step, producing gradually increasing contact forces. Fig.~\ref{fig:consistency} shows the results: despite identical inputs, the taxels of different sensors produce different outputs, confirming variability across sensors.\\
Finally, we mount the fingers on the ErgoCub hand and perform random 6D movements. We observe that tactile readings vary significantly with orientation. More critically, the displacement of the taxels follows gravity, producing opposite responses w.r.t. to contact cases.  For instance, when the fingertip is inverted, multiple taxels move outward due to gravity, a behavior that never occurs under actual contact.
\subsubsection{Data collection}\label{subsub:dataset}To estimate forces, we train a neural network to predict the 3D fingertip force vector from raw sensor readings. To collect data, we define contact points with Poisson sampling on the sensor surface, and for each point we generate random indentation trajectories. These are constrained within a cylinder centered at the contact point (radius $1~mm$ to prevent tangential slip, height $5~mm$ to avoid excessive forces). Data are collected by controlling the robot probe in position, ensuring contact across a wide variety of fingertip poses. During collection, the fingertip is fixed facing upward. To mitigate the gravity effect described in Sec.~\ref{subsub:characterization}, we augment the dataset with additional no-contact samples acquired by mounting the sensors on the robot and applying random wrist and finger motions, to avoid out-of-distribution inputs during experiments. In total, we collect about $200$k force–tactile pairs using all indenters, which are then balanced and reduced to $20$k per sensor. Results of the calibration are reported in Section~\ref{sub:experimentalresults}.

\subsection{Finger positions control}\label{subs:FingerPositionsControl}
We now describe how we control the fingers. Consider \(d\) fingers and let \(D_i \in \mathbb{R}^3,\ \dot{D_i} \in \mathbb{R}^3,\ \ddot{D}_i \in \mathbb{R}^3, \ i= 1,\dots,d\) be the position, linear velocity, and linear acceleration of the i-th finger, respectively.
Notice that the kinematic chain up to the hand palm is shared by all the fingers. Therefore, the joint positions vector for the i-th finger can be expressed as: \(q_i = [q_c^T\ q_{fi}^T]^T\), where \(q_c \in \mathbb{R}^{n_C}\) are the joints common to the finger kinematic chains (\(n_C=10\), that is torso-shoulder-elbow-wrist) and \(q_{fi} \in \mathbb{R}^{n_{Fi}}\) are those specific of the i-th finger (\(n_F = 2\) for thumb and index, \(n_F = 1\) for the others). We can construct a joint position vector:
\begin{figure}
	\centering
    \vspace{0.5em}
	\includegraphics[width=\columnwidth]{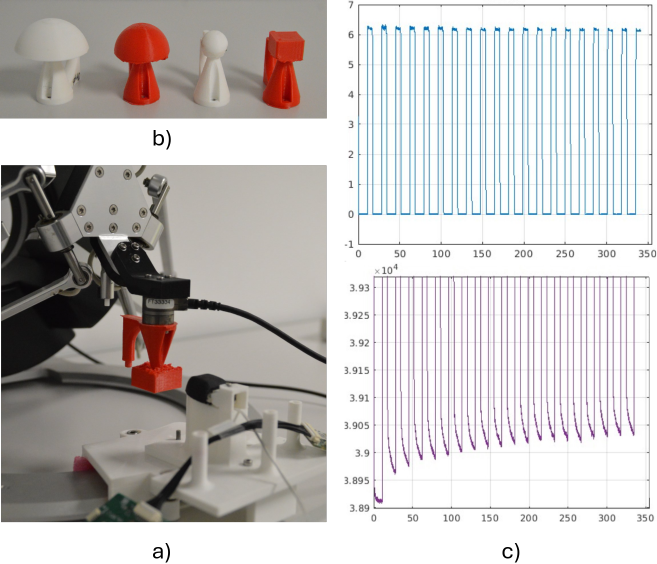}
	\caption{a) Data collection setup. b) Indenters. c) Effect of hysteresis on the tactile readings. The graphic above shows the indented force (N) over time (s), while the bottom graphic shows the output of the sensor's taxel over time. \label{fig:setup}}
    \vskip -2em
\end{figure}
\begin{equation}\label{eq:jointCompositionVec}
    q = 
    \begin{bmatrix}
        q_c \\
        q_{f1} \\
        \vdots \\
        q_{f_{d}}
    \end{bmatrix}
    \in \mathbb{R}^{n_C + n_{F1} + \dots  + n_{F_d}}
\end{equation}

The joint velocities \(\dot{q}\) and accelerations \(\ddot{q}\) are related to the finger linear acceleration as follows:
\begin{equation}\label{eq:FingerKin}
    \ddot{D}_i = \dot{J}_{P,i} \cdot \dot{q} + J_{P,i} \cdot \ddot{q}
\end{equation}
where \(J_{P,i} \in \mathbb{R}^{3 \times ({n_C + n_{F1} + \dots  + n_{F_d}})}\) is the positional part of the geometric Jacobian and \(\dot{J}_{P,i} \) its time derivative.
For each finger position, we are interested in calculating its orthonormal projection \(D_i^\Pi \in \mathbb{R}^3\) on the plane \( \Pi \):
\begin{subequations}\label{eq:FingerProj}
    \begin{align}
      & D_i^\Pi = O_{\Pi} + D_{t,i}^\Pi = D_i + D_{n,i}^\Pi \label{seq:FingerProj_1} \\
      & D_{t,i}^\Pi = (I_{3\times3} - P_n^{\Pi}) (D_i - O_{\Pi}) \label{seq:FingerProj_2}\\
      & D_{n,i}^\Pi = P_n^{\Pi} \cdot (O_{\Pi} - D_i) \label{seq:FingerProj_3}
    \end{align}
\end{subequations}

where \(P_n^{\Pi} \triangleq n \cdot n^T\) is the orthogonal projector onto the line spanned by \(n\); \(D_{t,i}^\Pi\) is the displacement, that lies on \( \Pi \), between the projection \(D_i^\Pi\) and the origin \(O_{\Pi}\); \(D_{n,i}^\Pi\) is the displacement, parallel to the unit normal vector \( n \), between the projection \(D_i^\Pi\) and the finger position \(D_i\).
Using the previously defined quantities, in the following we define the two modules composing the finger positions control: the QP problem, whose unknowns are the joint accelerations \(\ddot{q}\), and the \textit{Keep in Touch} module.
\subsubsection{\textbf{QP formulation}} we cast the finger positions control as:
\begin{equation}
    \begin{aligned}
& \underset{\ddot{q}}{\text{min}}
& & T_1 + T_2 + T_3 + T_4 \\
& \text{s.t.}
& & c_L \leq \ddot{q} \leq c_U
    \end{aligned}
\end{equation}
where the joint acceleration constraints are defined as (\cite{6301071}):
\begin{subequations}
    \begin{align}
      &c_U = \frac{k_l}{\delta t} \cdot (\frac{q^U - q}{\delta t} - \dot{q}) \\
      &c_L = \frac{k_l}{\delta t} \cdot (\frac{q^L - q}{\delta t} - \dot{q})
      \end{align}
\end{subequations}
with \(k_l\in(0,1]\), \(\delta t\) is the control period and \([q^L, q^U]\) is the joint range.
\subsubsection{\textbf{Bring the fingers on the plane}} we start by defining the terms of the QP. We want each fingertip position to lie on the plane \(\Pi\), so as to arrange the fingers on a planar support. We do so by imposing the following dynamics:
\begin{equation}\label{eq:FingerOnPlaneDyn}
    \ddot{D}_{n,i}^\Pi + K_{n,d} \cdot \dot{D}_{n,i}^\Pi + K_{n,p} \cdot D_{n,i}^\Pi = 0, \quad i= 1,\dots,d
\end{equation}
with \(\dot{D}_{n,i}^\Pi\) and \(\ddot{D}_{n,i}^\Pi\) first and second order time derivative of \eqref{seq:FingerProj_3}, \(K_{n,d}\) and \(K_{n,p}\) positive scalars. This is an asymptotically stable dynamics and converges to \(D_{n,i}^\Pi=0\), meaning that the i-th finger position converges to its projection on the plane \( \Pi \). Substituting \eqref{eq:FingerKin} and the second-order time derivative of \eqref{seq:FingerProj_3} in \eqref{eq:FingerOnPlaneDyn}, we obtain:
\begin{equation}
    A_{1,i} \cdot \ddot{q} = b_{1,i}, \quad i= 1,\dots,d
\end{equation}
with:
\begin{subequations}
    \begin{align}
      & A_{1,i} = P_n^{\Pi} \cdot J_{P,i} \\
      & b_{1,i} = K_{n,d} \cdot \dot{D}_{n,i}^\Pi + K_{n,p} \cdot D_{n,i}^\Pi + P_n^{\Pi} \cdot [\ddot{O}_{\Pi} - \dot{J}_{P,i} \cdot \dot{q}]
    \end{align}
\end{subequations}
for \(i= 1,\dots,d\). We define the relative cost function term as follows:
\begin{equation}\label{eq:FingerOnPlaneCost}
    T1 \triangleq \sum_{i=1}^{d} ||A_{1,i} \cdot \ddot{q} - b_{1,i}||^2_{W_1}
\end{equation}
where \(W_1>0\) and \(||x||^2_W = x^TWx\).
\subsubsection{\textbf{Align the centroid of the finger projections to the origin of the plane frame}}\label{Align the centroid} we want the position of the fingertips centroid projection to correspond to the origin of the plane frame, because this point is also the target for the CoP control (\ref{subs:PlanePose}). We define the error:
\begin{equation}\label{eq:FingersCenterError}
    e_{t} \triangleq O_{\Pi} - \frac{1}{d}\sum_{i=1}^{d} D_i^\Pi = - \frac{1}{d}\sum_{i=1}^{d} D_{t,i}^\Pi
\end{equation}
and impose the error dynamics:
\begin{equation}\label{eq:FingersCenterDyn}
    \ddot{e}_{t} + K_{t,d} \cdot \dot{e}_{t} + K_{t,p} \cdot e_{t} = 0
\end{equation}
with \(\dot{e}_{t}\) and \(\ddot{e}_{t}\) first and second order time derivative of \(e_{t}\), \(K_{t,d}\) and \(K_{t,p}\) positive scalars. The dynamics \eqref{eq:FingersCenterDyn} ensures that the geometric center of the  fingertips converges to the origin \(O_{\Pi}\). Substituting \eqref{eq:FingerKin} and the second-order time derivative of \eqref{seq:FingerProj_2} in \eqref{eq:FingersCenterDyn}, we obtain:
\begin{subequations}
    \begin{align}
      &A_2 \cdot \ddot{q} = b_2 \\
      &A_2 = (I_{3\times3} - P_n^{\Pi}) \frac{1}{d}\sum_{i=1}^{d} J_{P,i} \\
      \begin{split}
      &b_2 = \frac{1}{d}\sum_{i=1}^{d} [ \ddot{P}_n^{\Pi} \cdot (D_i - O_{\Pi}) + 2\dot{P}_n^{\Pi} \cdot (\dot{D}_i - \dot{O}_{\Pi})\\
        &\quad\quad+ (I_{3\times3} - P_n^{\Pi}) \cdot (\ddot{O}_{\Pi} - \dot{J}_{P,i} \cdot \dot{q})]\\
        &\quad\quad+ K_{t,d} \cdot \dot{e}_{t} + K_{t,p} \cdot e_{t}
      \end{split}
    \end{align}
\end{subequations}
and define the relative cost function term:
\begin{equation}\label{eq:FingersCenterCost}
    T2 \triangleq ||A_2 \cdot \ddot{q} - b_2||^2_{W_2}
\end{equation}
with \(W_2>0\).
\subsubsection{\textbf{Joint positions control}} we further add a postural term to keep the robot joint positions close to the initial configuration \(q^0\):
\begin{equation}\label{eq:FingerJointsCostRaw}
    - \ddot{q} - K_{d,q} \cdot \dot{q} + K_{p,q} \cdot (q - q^0) = 0 
\end{equation}
and  define:
\begin{equation}\label{eq:FingerJointsCost}
    T3 \triangleq \sum_{i=1}^{d} ||\ddot{q} - b_3||^2_{W_3} 
\end{equation}
with: \(b_3 = - K_{d,q} \cdot \dot{q} + K_{p,q} \cdot (q - q^0)\) and \(W_3>0\).
\subsubsection{\textbf{Bound Joint Accelerations}} this term bounds the required joint accelerations for the task and helps the solver convergence.
\begin{equation}\label{eq:BoundingCost}
    T4 \triangleq ||\ddot{q}||^2_{W_4} 
\end{equation}

\subsubsection{\textbf{Keep in touch}} in addition to the QP generated references \(q_{\Pi}\), we introduce a contribution \(\delta q_{\Pi}\) to compensate for small inaccuracies in the model, and mechanical imprecision due to finger tendon actuations. We call this module the \textit{Keep in Touch} controller. This contribution is added only to the finger joints \(q_{fi}\) and it is evaluated from the finger force \(F_i\). Given a force threshold \(F_{th}\), if \(F_{th} > F_i\),  then we update the offset so as to close the finger:  
\begin{equation}
    \begin{split}
        &\delta q_{\Pi} = min(max(K_{p,f} (F_{th}-F_i) + q_m^{fi} - \\
        & q^{fi}_{\Pi}, 0),q_u^{fi}-q^{fi}_{\Pi})
    \end{split}
\end{equation}
where \(q_u^{fi}\) is the upper bound for the range of \(q_{fi}\) and \(q_m^{fi}\) is the measured joint position, and \(K_{p,f}\) positive scalar.
On the other hand, when \(F_{th} < F_i\) and the finger is closing, if necessary, we only update this offset in order to guarantee that \(\delta q_{\Pi} < q_u^{fi} - q_{\Pi}\). The threshold is set to account for hysteresis effects.

\subsection{Plane pose evolution under the effect of the finger force measurements}\label{subs:PlanePose}

Based on the knowledge of the measured forces at the fingertips, we want to implement an \textbf{indirect} control of the CoP by adjusting the pose of the plane \(\Pi\). The idea is to move the CoP towards the plane frame origin, which is also the target position of the fingers centroid control (see \ref{Align the centroid}). We define \(F_i\) as the force measured by the i-th fingertip.

\subsubsection{\textbf{CoP calculation}} for each \(F_i\), we calculate the magnitude of the normal component \(R_{n,i}\in \mathbb{R}\) and the resultant \(R_n\in \mathbb{R}\) as:
\begin{subequations}
    \begin{align}
      &R_{n,i} \triangleq n^T \cdot F_i \\
      &R_n \triangleq \sum_{i=1}^{d} R_{n,i}
      \end{align}
\end{subequations}
Notice that \(R_{n,i} \geq 0\) because the fingertips can only push on the plane. Then, assuming that the force \(F_i\) is applied at the finger projection \(D^\Pi_i\), we calculate the CoP position relative to the origin \(O_{\Pi}\) as \cite{1325327}:
\begin{equation}\label{eq:CopToPlane}
    O_{\Pi}C = \sum_{i=1}^d \frac{R_{n,i}}{R_n} \cdot D_{t,i}^\Pi 
\end{equation}

\subsubsection{\textbf{Plane trajectory}}
We introduce the rotation matrix \(R_e\) as:
\begin{subequations}\label{eq:rotation_error}
    \begin{align}
      &R_e = Rotation(e) \label{rotation_error_1}\\
      &e = K_{p,e} s + K_{i,e} \int{s} \label{rotation_error_2}\\
      &s = O_{\Pi}C \times n \label{rotation_error_3}
    \end{align}
\end{subequations}
where \(Rotation(\cdot)\) calculates the rotation matrix from an axis-angle representation, \(K_{p,e}\) and \(K_{i,e}\) are positive scalars, and \(s\) is the cross product between \eqref{eq:CopToPlane} and the plane normal. The rotation matrix \(R_e\) is interpreted as the rotation that, applied to the initial plane orientation \(R_{\Pi}^i\), could bring the CoP towards the origin \(O_{\Pi}\).
Note that the integral action in \eqref{rotation_error_2} is introduced because the rotation due to the proportional action alone may be insufficient to cause the object to slide on the plane.
We generate the plane trajectory as in the following:
\begin{subequations}\label{eq:plane_trajectory_generation}
    \begin{align}
      &R_\Pi = R_e \cdot R^i_\Pi,\quad\omega_\Pi = 0,\quad\dot{\omega}_\Pi = 0 \label{plane_trajectory_generation_1}\\
      &O_\Pi = O^i_\Pi,\quad\dot{O}_\Pi = 0\quad\ddot{O}_\Pi = 0 \label{plane_trajectory_generation_2}
    \end{align}
\end{subequations}

with $O^i_\Pi$ is the (desired) initial plane position. 

\section{Experimental Results}
\label{sub:experimentalresults}
\begin{table*}[ht]
\vspace{0.5em}
\centering
\scriptsize
\caption{Results of the experiments. Force estimation uses  different networks for each finger (first five rows) and one network for all the fingers (last row). Each metric is evaluated at different positions (1–5) and on average. 
}
\label{tab:results_100_hz}
\begin{tabular}{
  l
  *{5}{S} S   
  *{5}{S} S   
  *{5}{S} S   
}
\toprule
\multicolumn{1}{c}{} &
\multicolumn{6}{c}{\textbf{Time (s)}} &
\multicolumn{6}{c}{\textbf{Distance CoP (cm)}} &
\multicolumn{6}{c}{\textbf{Success rate (\%)}} \\
\cmidrule(lr){2-7} \cmidrule(lr){8-13} \cmidrule(lr){14-19}
\multicolumn{1}{c}{\textit{Position}} &
\multicolumn{1}{c}{\textit{1}} &
\multicolumn{1}{c}{\textit{2}} &
\multicolumn{1}{c}{\textit{3}} &
\multicolumn{1}{c}{\textit{4}} &
\multicolumn{1}{c}{\textit{5}} &
\multicolumn{1}{c}{\textit{Mean}} &
\multicolumn{1}{c}{\textit{1}} &
\multicolumn{1}{c}{\textit{2}} &
\multicolumn{1}{c}{\textit{3}} &
\multicolumn{1}{c}{\textit{4}} &
\multicolumn{1}{c}{\textit{5}} &
\multicolumn{1}{c}{\textit{Mean}} &
\multicolumn{1}{c}{\textit{1}} &
\multicolumn{1}{c}{\textit{2}} &
\multicolumn{1}{c}{\textit{3}} &
\multicolumn{1}{c}{\textit{4}} &
\multicolumn{1}{c}{\textit{5}} &
\multicolumn{1}{c}{\textit{Mean}} \\
\midrule
\textbf{Object 1} & 9.2 & 6.5 & 6.4 & 42.1 & 40.5 & 20.9
                  & 35.0 & 13.2 & 13.6 & 33.4 & 31.9 & 25.4
                  & 66.7 & 100.0 & 100.0 & 100.0 & 100.0 & 93.3 \\
\textbf{Object 2} & 8.8 & 9.5 & 5.2 & 24.6 & 28.9 & 15.4
                  & 14.1& 6.4 & 3.1 & 12.6 & 15.0 & 10.2
                  & 100.0 & 100.0 & 66.7 & 100.0 & 66.7 & 86.7 \\
\textbf{Object 3} & 9.0 & 7.4 & 5.1 & 29.3 & / & 12.7
                  & 9.3 & 2.3 & 6.7 & 21.7& / & 10.0
                  & 100.0 & 100.0 & 100.0 & 66.7 & 0.0 & 73.3 \\
\textbf{Object 4} & 12.3 & 28.4 & 11.9 & 22.2 & 22.8 & 19.52
                  & 44.0 & 168.0 & 15.2 & 93.2 & 25.0 & 69.1
                  & 66.7 & 66.7 & 33.3 & 100.0 & 66.7 & 66.7 \\
\textbf{Object 5} & 42.3 & 20.94 & 39.7 & 5.7 & 43.0 & 30.3
                  & 34.6 & 21.7 & 10.3 & 21.0 & 33.2 & 24.2
                  & 66.7 & 100.0 & 100.0 & 100.0 & 100.0 & 93.3 \\
\bottomrule
`\textbf{Object 3 (SN)} & / & / & 26.8 & 5.1 & 11.7 & 11.5
                        & / & / & 5.8 & 1.2 & 2.7 & 2.9
                        & 0 & 0 & 67 & 100 & 67 & 47 \\
\bottomrule
\end{tabular}
\end{table*}

\begin{figure}
	\centering    \includegraphics[width=\columnwidth]{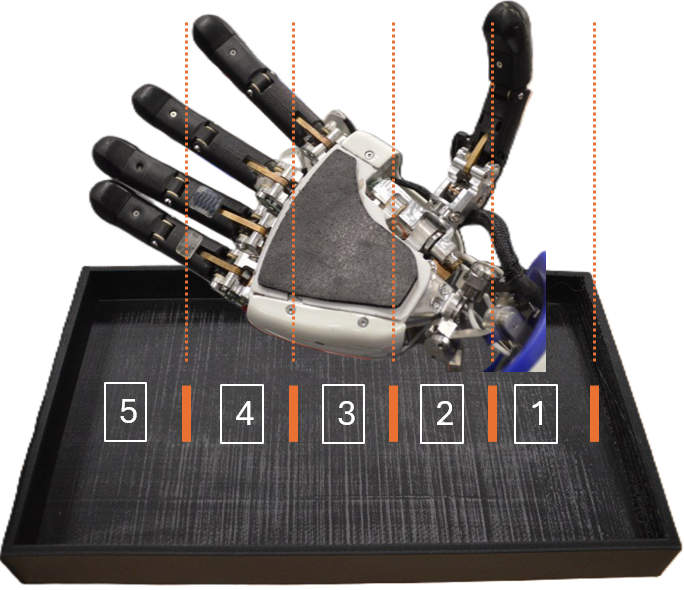}
	\caption{Starting position on the tray. The superimposed hand shows the position of the hand under the tray.\label{fig:position}}
    \vskip -2em
\end{figure}
We evaluate our method in two stages: first, by validating the precision of the force estimation network, and second, by testing the controller on balancing performance on a set of objects. The task consists in balancing an object on top of a 3D-printed tray held by the ergoCub hand. Because the hand is equipped with four Xela sensors, only four fingers are used. The tray weighs $250~g$, measures $27~cm$ x $17~cm$, and includes $2~cm$-high barriers ($0.5~cm$ thick) along the edges. These barriers prevent small objects from falling but still allow objects of higher mass to fall, ensuring the task remains non-trivial. The tray dimensions are constrained by the robot’s hand span, with the thumb–little finger distance being $25~cm$.
We test five objects spanning different shapes, materials, and internal dynamics:
\begin{itemize}
    \item Object 1–3: a paper tea box, filled with varying amounts of small plastic balls containing modeling clay of different weights ($122.7~g$, $152.6~g$, $216.2~g$), producing different mass distributions.
    \item Object 4: a fabric ball filled with sand ($98.6~g$).
    \item Object 5: an aluminum box filled with modeling clay ($125.6~g$).
\end{itemize}
Objects are chosen to weigh between $100~g$ and $300~g$, so that they could be reliably sensed by the Xela sensors, and  they remain below the tested fingertip limit of $5~N$ of sustainable load.

Each fingertip uses a dedicated force estimation network trained on its own calibration dataset to maximize accuracy. To assess the necessity of this choice, we conduct an ablation in which a single shared network is used across all fingers (evaluated on Object 3).
We perform all the experiments with the following parameters: $W_1=W_2=10.0$, $W_3=0.01$, $W_4=2.5$, $K_{p,q}=15.0$, $K_{d,q}=5.0$, $K_{t,p} = K_{n,p}=140.0$, $K_{t,d} = K_{n,d}=10.0$, $k_l=0.9$, $\delta_t = 0.01$, $K_{p,\omega} = 100$, $K_{p,e} = 8000* (\pi/180)$, $K_{i,e} = 25 * (\pi/180)$, $K_{p,f} = 50.0 * (\pi/180.0)$, $F_{th} = 0.15~N$. We limit the maximum plane orientation to \ang{30} along all three axes to constraint the speed of the object while moving on the tray. In the following we report results of the force estimation, of the balancing task employing single and multiple objects.

\subsubsection{Force estimation}\label{subsub:force_estimation}
We train a regressor using the dataset described in Sec.~\ref{subsub:dataset}. The regressor is a multilayer perceptron (MLP) with four linear layers, interleaved with LayerNorm and GELU activation functions. Following the characterization study in Sec.~\ref{subsub:characterization}, we address sensor-to-sensor variability by training a separate regressor for each sensor. We validate these regressors on test datasets collected by touching the sensors in poses different from those used for training. For comparison, we also train the same architecture on the combined data from all sensors and evaluate it on all test sets. On average, the single-sensor regressors achieve a normalized MAE of $9\%$, while the multi-sensor regressor reaches $13\%$.

\subsubsection{Single object balancing experiments}\label{subsub:single_obj} 
\begin{figure}
	\centering
	\vspace{0.5em}
    \includegraphics[width=\columnwidth]{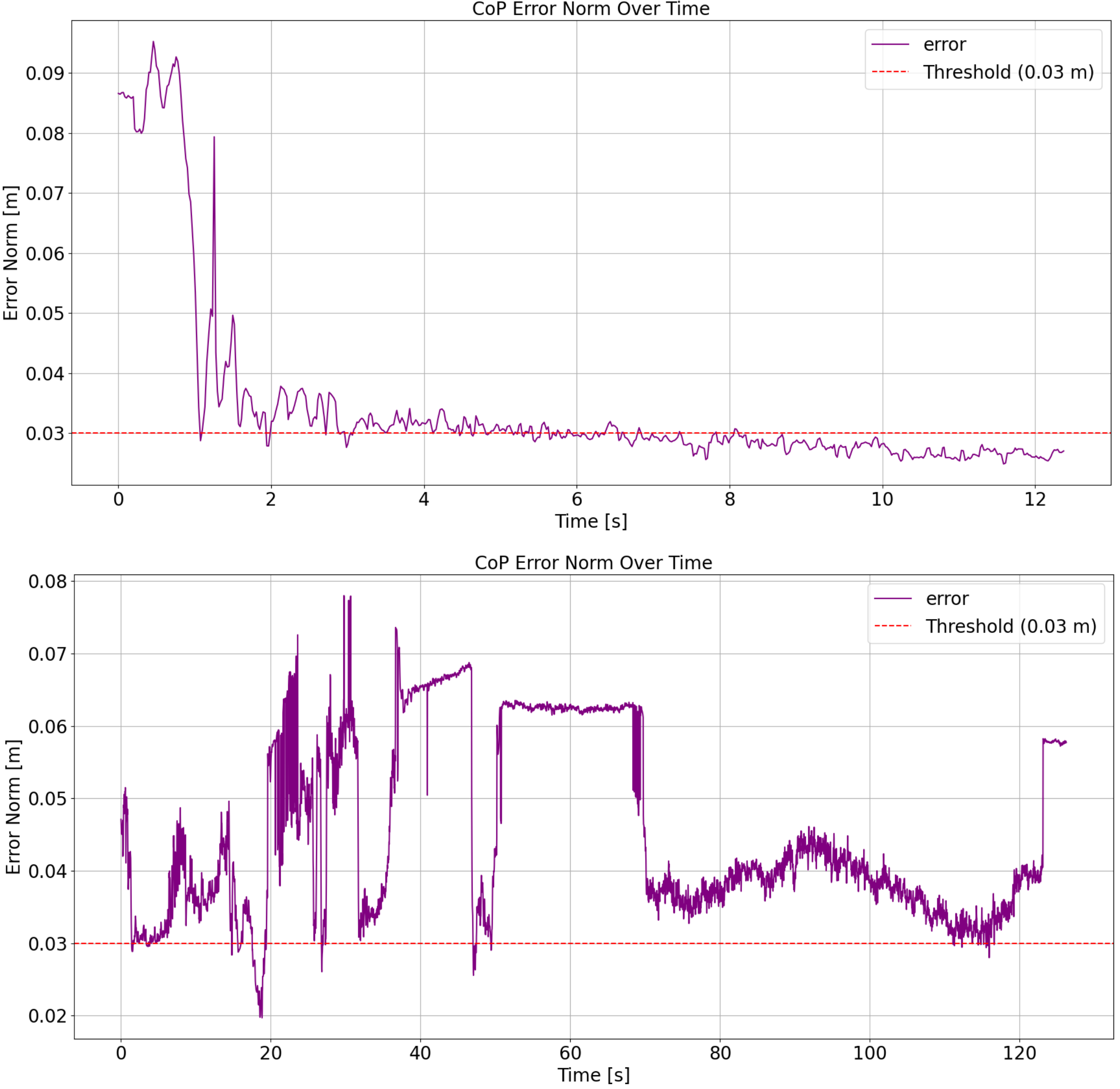}
	\caption{CoP error for a success (top) and a failure (bottom) episode. In the failure case, the error drops below the threshold but is highly unstable.\label{fig:error_time}}
    \vskip -2em
\end{figure}

For each object, we perform $15$ trials by placing it in five different positions on the tray, with three repetitions per position. The positions are chosen by uniformly dividing the tray along its major axis. Fig.~\ref{fig:position} shows the positions on the tray. We evaluate performance using three metrics. First, the time to convergence. Second, the distance traveled by the CoP, indicating how many corrections are required to balance the object (movements under $5~\text{mm}$ are ignored to filter sensor noise). Third, the success rate, when convergence is reached within $120~s$, also accounting for those cases in which the object falls. Converges is obtained by maintaining the distance between the CoP and the plate centroid under $3~\text{cm}$ threshold for at least $5~\text{s}$. This threshold ensures the CoP remains inside the fingertip contact polygon, while accounting for sensor noise and static friction effects, which prevent motions from small rotations of the plane, thus not allowing small corrections. Only successful runs are included in the Time and CoP Distance metrics. The first five rows of Table~\ref{tab:results_100_hz} summarize the results. The best performance is obtained with the lightest objects (Object 1 and 5), since their lower friction and small momentum when contacting the barriers are insufficient to lead to failure cases. In contrast, Objects 3 and 4 show the lowest performance. Balancing Object 3 is difficult because higher friction requires larger plane inclinations to induce motion, often leading to unrecoverable situations. Object 4, instead, can roll easily; even small rotations generate high momentum, causing strong impacts against the barriers and eventual falls. Regarding object position, \textit{Position 4} is the easiest to balance, being closest to the center of the support polygon (around the index/middle finger), whereas \textit{Positions 1, 2, and 3} are farther away and therefore harder to control. Small plane inclinations, needed to correct minor unbalances, strongly affect Object 4, while Object 3 demands larger inclinations due to static friction. When the CoP is near the support polygon center, the controller also reacts more slowly. This behavior is reflected in the \textit{CoP Distance} metric: Object 4 exhibits much larger CoP movements, while Object 3 remains comparable to the others. Finally, to validate our choice of training a separate force estimator for each finger, we conduct an ablation using a single shared network (\textbf{SN}) on Object 3, shown in the last row of Table~\ref{tab:results_100_hz}. Results show a $26\%$ performance drop, confirming the benefit of finger-specific estimators: when force estimation is unreliable, the system struggles to converge. This outcome is consistent with the findings in Section~\ref{subsub:force_estimation}. \\
Next, we analyze the primary reasons for failure for each object. Object 3 and 4 consistently fail because of strong momentum that impacts the barrier. Object 2 fails once because of the momentum and once because of exceeding convergence time, Object 5 for strong impact with barriers; in the $SN$ case, Object 3 fails seven times because of exceeding convergence time, while one time it fails because its speed generates a large impact with the border of the table. In general, potential inaccuracies in the force estimation may affect the outcome of the experiment. The results are expected, as the heavier object (Object 3) and the object with less static friction (Object 5) tends to accumulate more momentum. Fig. \ref{fig:error_time} shows the CoP error over time for a failure and for a success episode. The accompanying video provides a visual demonstration of the objects and experiments.
\subsubsection{Multiple objects balancing experiments}\label{subsub:multiple_obj}
To further check the robustness of our approach, we conduct experiments with multiple objects simultaneously. In particular, we use two aluminum boxes filled with modeling clay (as in Object 5), for a total weight of $299~g$. We randomly place the boxes on the tray, varying their position. The boxes are placed in three predefined configurations (opposite sides, same side, and center), with exact positions randomized across five trials each, obtaining 80\%, 60\%, and 100\% success rate respectively. All failures result from excessive plane tilting caused by strong impacts with the objects. In successful trials, the average convergence time is $11.3~s$, and the CoP travels $13~cm$ on average. These outcomes are consistent with the single-object experiments reported in Section~\ref{subsub:single_obj}. When the objects are placed on the same side, the CoP error increases, requiring larger plane inclinations; this amplifies object momentum and leads to strong barrier impacts. Such failures occur primarily when the objects move simultaneously, behaving as a single mass. Failures with objects set apart arise for a similar reason: the plane inclines to reduce the error, one object slides rapidly toward the other, and the impact with the plane border causes failure. These outcomes are expected, as they mirror the dynamics observed in the single-object case.

\subsubsection{Ablation on control frequency}\label{subsub:ctrl_freq}
To assess the importance of the controller reactivity when dealing with object with varying mass, we repeat the same tests for single objects reported in Section \ref{subsub:single_obj}, lowering the control frequency to $50~Hz$. The lower frequency reduces the system reactivity and hinders proper recovery, especially for fast-moving or rolling objects (e.g., Objects 3 and 4): on average, we experience a $9\%$ drop in success rate. 

\section{Limitations}
In this section, we discuss the limitations of our work. The first concerns discrepancies between the 3D hand model and tendon friction, which cause misalignment between desired and actual finger motion. Nevertheless, the approach is still able to drive the CoP in most cases, showing robustness to modeling errors. Furthermore, the \textit{Keep in Touch} control is currently implemented as a separate module: a more systematic approach would integrate this compensation directly into the QP.
The second limitation is the plane trajectory module: the trajectory generation in \eqref{eq:plane_trajectory_generation} only accounts for the angular position of the plane $\Pi$, while its linear position remains constant. Exploiting the full pose of the plane could enhance the balancing performance and also enable the execution of more complex tasks, such as trajectory tracking with the arm while balancing an object.
Finally, contact with the palm represent an additional source of error: in some trials, the plane tilts and touches the unsensorized palm, breaking the assumption of fully sensorized contact surfaces. A sensorized palm would mitigate this issue. 

\section{Conclusion}
In this paper, we presented a model-based force-aware control strategy for multifingered robotic hands. By operating in the force domain, our method adapts the motion of the torso, arm, wrist, and fingers to maintain stable contact with objects of varying mass, under the assumption of planar contact; we validated the approach on single-object and multi-object balancing tasks involving diverse shapes and weights, demonstrating robust performance across all scenarios.
As future works, we  plan to integrate the proposed algorithm as a low-level policy within hierarchical manipulation frameworks, particularly for tasks involving bimanual manipulation.

\bibliographystyle{IEEEtran}
\bibliography{bibliography}

\end{document}